# Monocular Vision based Collaborative Localization for Micro Aerial Vehicle Swarms

Sai Vemprala and Srikanth Saripalli[1]

*Abstract*— In this paper, we present a vision based collaborative localization framework for groups of micro aerial vehicles (MAV). The vehicles are each assumed to be equipped with a forward-facing monocular camera, and to be capable of communicating with each other. This collaborative localization approach is built upon a distributed algorithm where individual and relative pose estimation techniques are combined for the group to localize against surrounding environments. The MAVs initially detect and match salient features between each other to create a sparse reconstruction of the observed environment, which acts as a global map. Once a map is available, each MAV performs feature detection and tracking with a robust outlier rejection process to estimate its own six degree-of-freedom pose. Occasionally, the MAVs can also fuse relative measurements with individual measurements through feature matching and multiple-view geometry based relative pose computation. We present the implementation of this algorithm for MAVs and environments simulated within Microsoft AirSim, and discuss the results and the advantages of collaborative localization.

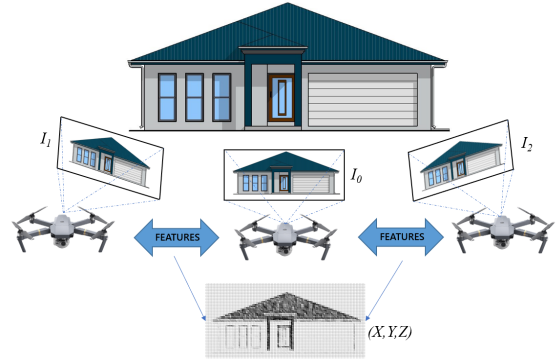

**Fig. 1:** Collaboration between multiple MAVs to match feature points and to obtain a sparse reconstruction of the environment ahead. This reconstruction is then shared as a global map between all members of a group.

## I. INTRODUCTION

Micro aerial vehicles (MAV) are currently becoming the platforms of choice for various robotic applications. Their agility and ability to navigate within remote or cluttered spaces has created heightened interest in their usage for missions such as 3D mapping, search and rescue and agricultural monitoring. In order to adhere to their on-board requirements, small and low-power sensors which can still produce rich information are deemed suitable for MAV localization and navigation, and in this regard, monocular cameras have shown great potential. Cameras are also suitable for areas with no a-priori environmental information, or GPS-denied areas. At the same time, given the small size of MAVs and thereby, low payload carrying capacity and on-board computational power, collaboration between the vehicles has the potential to boost mission efficiency. Collaboration allows distribution of tasks, while also giving the ability to offload computationally heavy tasks to perhaps a leader MAV, or to a ground station. At the same time, collaboration can also help enhance the accuracy of localization as multiple sources of information can be fused for robust estimation. This is especially useful in the case of monocular vision sensing, as the scale factor is impossible to determine from a single camera without additional information, and with the possibility of collaboration, it can be obtained using information from other MAVs in a group.

In this paper, we present an approach for vision based collaborative localization (VCL) for a group of MAV, as an extension to our preliminary work in this area presented in [1], [2]. We assume that each MAV is equipped with a forward facing monocular camera, and is capable of communicating with the other MAVs in order to share map information and to transmit or receive relative pose data. Our framework is distributed in nature and hence would not require constant communication between vehicles. In the first step of the algorithm, the MAVs perform feature detection and matching using the images captured from their on-board cameras. These common features are then triangulated to form a sparse reconstruction, and a global map is created that all vehicles have access to (figure 1). Once the MAVs start moving, each MAV performs its own individual pose estimation by tracking features from the global map, which we call intra-MAV localization. When required, one MAV can also generate a relative measurement to a target MAV, which is then fused with the target's own estimated state using the covariance intersection technique: and this process is termed inter-MAV localization (figure 2). As the MAVs continue to navigate, if the number of tracked features for the MAVs consistently falls below a threshold, the MAVs match the visible features from each camera again to update the global map. This paper presents improvements to our previous approach and adds the inter-MAV localization part to the localization algorithm, whereas previously the aspect of collaboration was only for updating maps.

## II. RELATED WORK

Vision based localization has been studied extensively in the literature. Monocular SLAM have been investigated on-board multirotor vehicles, which try to remove scale ambiguity either by fusing vision data with an IMU [3], by using multiple initial views [4] to obtain metric scale information, or by directly using dense photometric information [5].

[1]Sai Vemprala and Srikanth Saripalli are with the Department of Mechanical Engineering, Texas A&M University, College Station, Texas, USA svemprala@tamu.edu, ssaripalli@tamu.edu

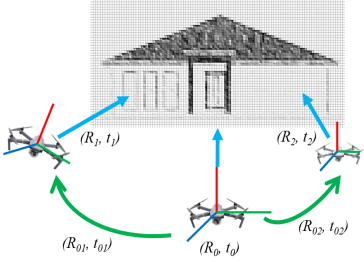

**Fig. 2:** Inter-MAV and intra-MAV localization: a visual representation. Each drone computes its own pose (intra-MAV), represented as 3-axis positions, but they also have the capability to compute relative pose to a neighboring drone (green arrows).

There has been significant amount of work done on the theoretical foundations of collaborative localization over the past decade. Martinelli et al. [6] present a multi robot localization approach that fuses proprioceptive and exteroceptive measurements using an extended Kalman filter. Nerurkar et al [7] present a maximum a posteriori cooperative localization algorithm in a distributed fashion, through continuous and synchronous communication within the robot group. In [8], the authors present a fully decentralized cooperative localization approach where the robots need to communicate only during the presence of relative measurements, an algorithm we use in our paper to facilitate inter-MAV data fusion. Indelman et al [9] propose a multi robot localization algorithm that can handle unknown initial poses and solves the data association problem through expectation maximization. Knuth and Barooah [10] propose a distributed algorithm for GPS-denied scenarios, where the robots fuse each other's information and average the relative pose data in order to achieve cooperative estimation.

Only very recently have there been advances in the realm of collaborative localization using vision and aerial vehicles. Indelman et al. [11] propose a three-view geometry inspired technique for performing cooperative localization for camera-equipped vehicles. Zou and Tan [12] present a collaborative monocular SLAM system with a focus on handling dynamic environments, with multiple vehicles helping each other isolate moving features from constant ones. In [13], the authors present an approach where two UAVs equipped with monocular cameras and IMUs estimate relative poses along with absolute scale, thus acting as a collaborative stereo camera. Piasco et al. [14] also present a distributed stereo system with multiple UAVs for collaborative localization, with a focus on achieving formation control. Forster et al. [15] show a collaborative SLAM system based on a structure from motion pipeline with a centralized ground station merging maps based on overlap: but the vehicles do not receive any additional information from the central server or other vehicles. Schmuck and Chli [16] present a collaborative monocular SLAM pipeline with a centralized paradigm where each MAV runs the ORB-SLAM2 pipeline and a central server focuses on place recognition, optimization and map fusion.

In our paper, we describe a collaborative approach focused mainly on localization that combines individual and relative estimation for MAVs. Exploiting multiple view geometry techniques and individual pose estimation enables the MAVs to move freely in an unconstrained way, and the distributed nature of the algorithm makes sure that the MAVs need to communicate only when a relative pose correction is needed, or when the map needs to be updated. During the intra-MAV estimation step, each MAV needs to perform only feature tracking and pose estimation, which greatly reduces the computational load compared to a full monocular SLAM pipeline. One example application of a pipeline such as this would to create a decoupled aerial imaging system: where multiple MAVs collaborate to simulate a variable baseline stereo imaging system, thus being able to map large 3D structures or areas efficiently. This localization pipeline also forms the base for our current work on uncertainty-aware planning for swarms of MAVs [17].

## III. PROBLEM DEFINITION

Consider a group of $N$ MAVs indexed as $m = 1...N$. At each instant of time, denoted in discrete counts as $k = 0, 1, 2, ...$, each MAV obtains an image of the environment ahead of it using its monocular camera. At time $k = 0$, in our implementation, one of the vehicles is chosen as the center of the group, and its initial location is fixed as the origin of the global Cartesian coordinate frame. From then on, all vehicle positions are estimated with respect to this initial frame. This global origin can be chosen as desired: for instance, it could be set to a world landmark or a significant location; or in cases of formation control, the leader MAV's location at every instant could be chosen as the origin against which the other vehicles are localized.

At the beginning, two or more MAVs participate in a map building step: where common features are isolated and used to obtain a sparse reconstruction of the surrounding environment. This reconstruction is then shared between all MAVs as a global map. At every time step $k$, each MAV has access to an image captured on-board and needs to estimate a 6 DoF 'pose', which is expressed in the coordinate frame as selected above. The pose of the $i$-th MAV at time step $k$ thus becomes part of a projective transformation matrix for that MAV's camera as:

$$m_k^i = K_i \begin{bmatrix} \mathbf{R} & \mathbf{t} \\ 0 & 1 \end{bmatrix} \quad (1)$$

where $\mathbf{R} \in SO(3)$ and $\mathbf{t} \in \mathbb{R}^3$ and $K$ denotes the intrinsics matrix. The pose thus computed is encoded in the state of each vehicle, which can be written for vehicle $m$ at time step $k$ as (expressing rotations as Euler angles):

$$\mathbf{x}_k^m = \begin{bmatrix} (\mathbf{t}_k^i)^\top & \phi & \theta & \psi \end{bmatrix}^\top \quad (2)$$

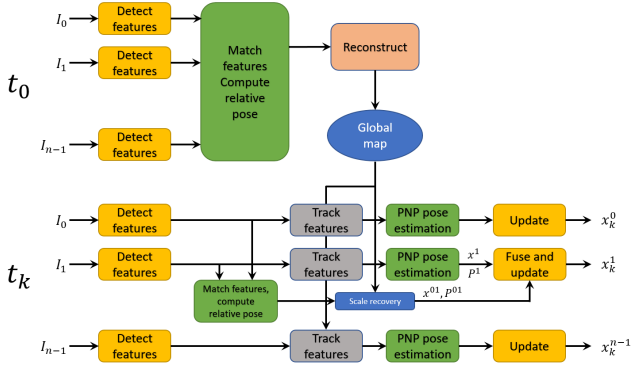

**Fig. 3:** Flow of the vision based collaborative localization (VCL) algorithm. At any time step $k$, the MAVs are able to fuse individual (intra-MAV) estimation and relative (inter-MAV) estimation to produce a final pose estimate.

The aspect of collaboration comes into picture when the MAVs need to communicate with each other in order to build or update a 3D map, or when one of the MAVs requests a relative measurement from another. In the latter case, the requesting MAV is treated as a 'target' MAV, and the MAV responding to this request then computes a relative pose measurement. The target MAV then receives this measurement and fuses the information with its own local estimate to produce a possibly more accurate pose. This localized correction procedure is more advantageous than a full map update because only two MAVs need to participate in this step, unlike the map update, which would have to be communicated to all the vehicles. In case the MAVs move far away from the initial map, or if the map is changing, the number of tracked features goes low: and if it falls below a certain threshold for all MAVs, the map is updated through another reconstruction.

## IV. VISION BASED COLLABORATIVE LOCALIZATION

In this section, we present a detailed discussion of the various steps of the collaborative localization algorithm. A high level chart of the algorithm flow is shown in figure 3. Initially, we list the assumptions made:

1) All the cameras on-board the MAVs are calibrated and the camera intrinsics are known.
2) The distance between any two MAVs at the beginning is known.
3) The MAVs are capable of communicating feature and pose data between each other, and in this presentation of our work, we ignore delays in communication.

### A. Feature detection and matching

In order to proceed with pose estimation, the first step would be to identify salient features in the surrounding environment from the cameras. Subsequently, these features can be matched between multiple views, which results in a set of common features that can be used to estimate the transformation between the views.

In our algorithm, we utilize accelerated KAZE (AKAZE) features. AKAZE features are multi-scale features which are faster to compute compared to SIFT/SURF and demonstrate better accuracy than methods such as ORB [18]. The low computational necessity is also in part due to the utilization of modified local difference binary (M-LDB) descriptors. We currently utilize brute force matching on a Hamming distance metric in order to match these binary descriptors and result in feature matches. Although these are the currently used techniques, our localization algorithm is method-agnostic and can be adapted to any type of feature extraction or matching.

### B. Relative pose estimation

Once common features are computed and isolated as described above, relative pose estimation between the MAVs is one of the integral steps in the collaborative localization algorithm. This relative estimation can be used for multiple tasks: to reconstruct and update the global map, to correct the pose of another MAV and so on.

Given two sets of corresponding feature points observed on the image plane, the transformation between these two views is encoded in the essential matrix. We utilize the five-point algorithm [19] to compute the essential matrix $E$. The relative rotation $R$ and translation $t$ between two views are represented within the essential matrix as:

$$E = [t]_\times R \qquad (3)$$

We here note that the precision of the essential matrix depends heavily on the fidelity of the feature matches obtained as described in subsection A. When navigating in sparsely populated environments, at high speed, or when observing repetitive feature data (which is common for textures in urban settings), feature matching is susceptible to a great amount of inaccuracy, resulting in false matches, which can then affect the relative pose estimation. A conventional way of solving this problem is by using the random sample consensus method (RANSAC): an iterative method that seeks to find outliers from the provided set of feature matches.

RANSAC typically requires the choice of a parameter known as threshold ($T$), which determines the confidence. But as in our application, we would require the pose estimation to happen multiple times while the vehicles are in motion, the noise levels of the images/features would not be constant, and presetting the threshold parameter could result in degradation of performance over time. To avoid this problem, we create a more robust approach for relative pose estimation by utilizing a technique known as a-contrario RANSAC (AC-RANSAC) method, proposed in [20]. This a-contrario approach adaptively chooses a choice of the parameter $T$ according to the noise in the given data, through which we propagate sets of feature matches. The essential matrix $E$ also encodes the information about how the image points are related to the epipolar lines in the other image: a property that's exploited in order to remove outliers. Feature matches which are 'farther' from the epipolar lines as per the current consensus are considered to be inaccurate and discarded.

## C. Map building

Once a relative pose is obtained between two views, this information can be used to compute a sparse reconstruction of the surrounding environment. We use the Hartley-Sturm triangulation method [21] to obtain a 3D map of the matched points according the relative pose computed as described in the previous subsection. This map is then made available to all the MAVs and can be reused for feature tracking, 3D-2D correspondence computing as well as a source for performing scale factor recovery for subsequent reconstructions. The globally available map data consists of the 3D locations of all the points, as well as the feature descriptor data associated with each point. We note here that in the beginning, according to assumption 2, having access to the distance between two MAVs helps remove the scale ambiguity problem with the reconstruction. For a group containing more than two vehicles, all MAVs capture images from their cameras, and a visibility graph of feature overlap is generated in order to isolate common features. The pair with the highest number of overlapping features is considered a seed pair and a first reconstruction is attempted. Once a reconstruction is generated, other MAVs are incrementally included in this reconstruction based on the features being tracked by them. Finally, we refine the poses and scene using fast bundle adjustment.

## D. Intra-MAV localization

We define intra-MAV localization as the step performed on-board each MAV in the group where the MAV tries to estimate its own pose individually. Once a global map is available, for every new image captured at a time step, each MAV attempts to track the features from the global map that are still visible in the current view, thus obtaining a set of tracks which are correspondences between 2D points observed in the image matched with 3D points in the existing map. The intra-MAV localization involves applying the perspective-N-point algorithm [22] to these correspondences in conjunction with another AC-RANSAC scheme, in order to estimate the position and orientation of the MAV. Once a pose estimate is computed, we attempt to refine the pose by minimizing the reprojection error as defined below:

$$m^* = \arg\min_{m} \sum_{i} \|x_i - P(X_i, m)\| \quad (4)$$

$P$ encodes the camera projective transformation of a 3D point $X_i$ onto the image plane for the computed pose $m$, whereas $x_i$ is the set of actual feature point locations in the image at that time step. Through this step, we also obtain the solution quality encoded within a covariance matrix, which, combined with the final reprojection error, we then use to scale the measurement noise covariance for that MAV. We discuss this uncertainty estimation in more detail later in the paper.

At each time step, the MAV's pose is predicted using a Kalman filter framework. Once a pose is computed, the obtained measurement is used to correct the state and covariance of that particular MAV. If a measurement obtained at time step $k$ by MAV $i$ is denoted as $z_k^i$ with a covariance matrix $R_k^i$, the measurement and its noise covariance are then used to correct its own pose at time step $k$.

$$\mathbf{K}_k = \mathbf{P}_{k|k-1}\mathbf{H}_k^\top(\mathbf{H}_k\mathbf{P}_{k|k-1}\mathbf{H}_k^\top + \mathbf{R}_k)^{-1} \quad (5)$$
$$\mathbf{x}_{k|k}^i = \mathbf{x}_{k|k-1}^i + \mathbf{K}_k(\mathbf{z}_k - h(\mathbf{x}_{k|k-1}^i)) \quad (6)$$
$$\mathbf{P}_{k|k}^i = (\mathbf{I} - \mathbf{K}_k\mathbf{H}_k)\mathbf{P}_{k|k-1} \quad (7)$$

## E. Inter-MAV localization

The inter-MAV localization step is an attempt by one MAV to estimate the location of another through a relative measurement. Typically useful when one MAV is able to localize better (owing to better feature visibility, proximity to texture-rich areas etc.), that MAV computes the relative pose to another, and the latter subsequently fuses this relative estimate with its on-board individual estimate. During this step, feature matches between two MAVs are isolated and used to compute a relative pose. While this results in a relative measurement, this has an arbitrary scale factor. In order to compute the right scale factor $\lambda$, we consider the 'host' MAV to be at the origin and the 'target' MAV located at $[R|t]$, where $R$ and $t$ are the estimated relative rotation and translation. Once a reconstructed local map is available, this map is compared with the global map to isolate common 3D points, and thereby recover the scale factor. We here recall the fact that the global map is considered metrically accurate due to the presence of information about distance between at least two MAVs at the first time step. Once this scale factor is computed, the relative pose is scaled accordingly, and refined similar to the intra-MAV estimation step: through minimization of the reprojection error.

Intra-MAV estimation usually suffers in accuracy when there are not enough features to be tracked from the original map or if the feature locations form a degenerate case. In such cases, inter-MAV estimation can be helpful as it utilizes common features between the MAVs at that instant and does not require multiple observations over time, and because relative pose estimation tends to be more robust. At the same time, scale recovery in the inter-MAV estimation step requires a minimum of only two matches between the local and global map, as opposed to intra-MAV estimation, which requires a significantly higher number of tracked features for better accuracy.

After the relative measurement is computed, there is a newly obtained piece of information, namely, the relative pose from MAV $i$ to MAV $j$ which needs to be fused with the local information of MAV $j$. Yet, it should be noted that this relative measurement and the measurement that resulted in the local pose of MAV $j$ have common sources of information, i.e., the map data and the features. At the same time, these two MAVs could have communicated pose data in the past, which makes these estimates correlated. But because these cross-correlation parameters are not kept track of, the correlations are treated as unknown: which makes the conventional EKF update step result in inconsistent and erroneous estimates. Hence, we utilize the approach described in [8] to fuse these estimates using covariance intersection. Covariance intersection, first proposed in [23], was an approach developed to consistently fuse estimates that have unknown correlations. In order to perform this fusion, first, MAV $i$ computes its own estimate of the state and covariance for MAV $j$ using the relative measurement $z_k^{i,j}$, which we denote as $x_k^{j'}$ and $\mathbf{P}_k^{j'}$.

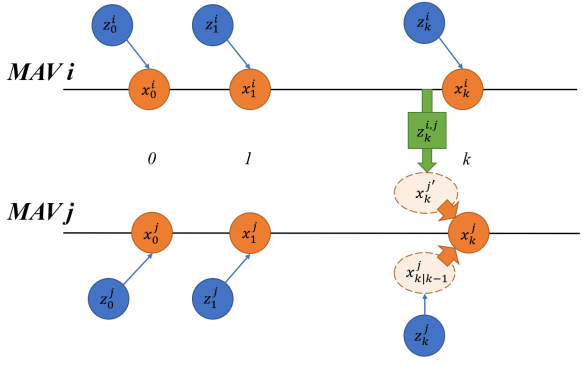

**Fig. 4:** A pictorial representation of inter-MAV localization. At time step $k$, MAV $i$ attempts to correct the pose of MAV $j$ by generating a relative measurement, which is then fused by MAV $j$ with its own on-board estimate.

$$\mathbf{x}_k^{j'} = \mathbf{x}_{k|k-1}^i + \mathbf{M}_k^{i,j} \mathbf{z}_k^{i,j} \qquad (8)$$

$$\mathbf{P}_k^{j'} = \mathbf{H}_k^{i,j} \mathbf{P}_{k|k-1}^i \mathbf{H}_k^{i,j\top} + \mathbf{M}_k^{i,j} \mathbf{R}_k^{i,j} \mathbf{M}_k^{i,j\top} \qquad (9)$$

At the same time, it is assumed that MAV $j$ has already performed its own step of intra-MAV estimation and hence has a locally computed state-covariance pair. Then the CI algorithm is used to fuse these two estimates as below.

$$\mathbf{P}_{k|k}^j = \left[\boldsymbol{\omega}(\mathbf{P}_{k|k-1}^j)^{-1} + (1-\boldsymbol{\omega})(\mathbf{P}_k^{j'})^{-1}\right]^{-1} \qquad (10)$$

$$\mathbf{x}_{k|k}^j = \mathbf{P}_{k|k}^j \left[\boldsymbol{\omega}(\mathbf{P}_{k|k-1}^j)^{-1}\mathbf{x}_{k|k-1}^j + (1-\boldsymbol{\omega})(\mathbf{P}_k^{j'})^{-1}\mathbf{x}_k^{j'}\right] \qquad (11)$$

Where $\omega$ is a parameter that is computed such that the trace of the resultant covariance matrix is minimized.

*F. Uncertainty estimation*

One of the critical parts of localization is to estimate not only the position and orientation of the vehicle, but also estimate the uncertainty of the estimated pose, which is usually described through the pose covariance. This pose covariance is conventionally propagated through a Kalman filter framework while predicting and updating poses. In our algorithm, we also follow a Kalman Filter predict-update procedure; but in order to accurately model the accuracy of measurements, it is essential to represent the accuracy of the pose estimation itself within the filtering scheme by obtaining an estimate of the uncertainty on the measurement.

In both the inter and intra-MAV estimation steps, the final refinement of the pose is performed through a non-linear least squares method where the algorithm attempts to minimize the reprojection error. Within this algorithm, a Jacobian matrix is computed between the changes in reprojection error seen for changes in the pose. The outer product of this Jacobian matrix at the final optimum with itself is an approximation of the Hessian matrix of the solution, while the inverse of this Hessian matrix is an approximation of the covariance matrix of the reprojection errors [24]. Hence, the approximate covariance of the solution can be expressed as

$$\Sigma = (J^\top J)^{-1} \qquad (12)$$

It is important to note here that $\Sigma$ does not necessarily translate into pose covariance: it merely expresses the quality of the solution and the possible uncertainty around the local surface at the point of convergence. In case the solution is a local minimum, the estimated covariance could still be low while the pose estimate is far from the actual value. So in order to express the pose uncertainty more accurately, we scale this covariance with the actual reprojection error obtained for that pose estimate and use this scaled value as an estimate of the measurement noise covariance in the update step of the Kalman filter.

$$R = (J^\top J)^{-1} * \epsilon_{rms} \qquad (13)$$

---

**Algorithm 1** Inter-MAV, intra-MAV localization procedures

1: **procedure** LOCALIZEINTERMAV($x_i, I_i, I_j$)
2:     $p_i, p_j \leftarrow detectFeatures(I_i, I_j)$
3:     $\bar{p}_i, \bar{p}_j \leftarrow matchFeatures(p_i, p_j)$
4:     $E \leftarrow \text{ACRANSAC}(\bar{f}_1, \bar{f}_2, K_i, K_j)$
5:     $R, t \leftarrow svd(E)$
6:     $P_1, P_2 \leftarrow [I|0], [R|t]$
7:     $O' \leftarrow reconstruct(\bar{f}_1, \bar{f}_2, P_1, P_2)$
8:     $M_{map} \leftarrow matchFeatures(O', O)$
9:                         ▷ $O'$: Local map, $O :=$ Global map
10:     $\lambda \leftarrow recoverScale(M_{map})$
11:     $[R, t] \leftarrow [R, t] * \lambda$
12:     $z_k^{i,j}, R_k^{i,j} \leftarrow refinePose(R, t, O)$
13:     $x_k^{j'}, P_k^{j'} \leftarrow eqn(8), (9)$
14:     **return** $x_k^{j'}, P_k^{j'}$
15:
16: **procedure** LOCALIZEINTRAMAV($I_k^j, K, O$)
17:     $x_{k|k-1}^j, P_{k|k-1}^j \leftarrow predictState()$
18:     $k \leftarrow detectFeatures(I_k^j)$
19:     $\bar{p} \leftarrow trackFeatures(p_k, O)$
20:     $R, t \leftarrow PNP(\bar{p}, O, K)$
21:     $z_k^j, R_k^j \leftarrow refinePose(R, t, O)$
22:     $x_k^j, P_k^j \leftarrow updateState(z_k, R_k, x_{k|k-1}^j, P_{k|k-1}^j)$
23:                                 ▷ $eqn(6), (7))$
24:     **return** $x_k^j, P_k^j$
25:

*G. Map updates*

In cases where the number of features tracked by a majority of the MAVs drops below a certain threshold, a map update is performed through collaboration. This process is similar to the reconstruction attempted at the first time step. As this map update is performed while some points from the previous map are still visible from the MAVs, common features between the old and new maps are compared in order to recover the scale of the new reconstruction. This new map is then re-utilized as the global map.

## V. Implementation and Results

In this section, we present details about our implementation and preliminary results obtained using the vision based collaborative localization (VCL) algorithm. The algorithm was tested in a simulation setting, with Microsoft AirSim as the primary simulator used for flight and image capture. AirSim [25] is a MAV simulator that uses Unreal Engine as its base, thus being able to simulate realistic scenes, shadows, post-processing etc. In computer vision based applications, using a high fidelity graphical environment such as ones that can be generated using Unreal Engine enables testing the algorithms in close-to-real-life settings. Each MAV simulated within AirSim had a forward facing monocular camera, and on-board images of 640x480 resolution were captured at approximately 5 Hz. These images and ground truth were recorded and the VCL algorithm was tested offline on this pre-recorded data. We utilize open-source libraries available in OpenCV [26] and OpenMVG [27] in order to implement feature detection, matching, AC-RANSAC and PNP pose estimation. Ceres libraries [28] were utilized to refine reconstructions and estimated poses, as well for estimating covariances of the solutions.

### A. Inter-MAV localization

One of the problems that is evident in single monocular-camera localization is the issue with pure rotation movement in the yaw direction, which is a very common maneuver for MAVs. Pure rotation usually causes existing map points to go out of view, while the fact that there is no translation by the camera means it is not possible to triangulate new feature points through a single camera without any additional information. In contrast, the VCL algorithm is able to handle this problem by matching feature points between what is rotating MAV and another MAV that is observing common scene points.

As a test case, we simulate an environment containing two perpendicular buildings being observed by an MAV (MAV 0) that performs periodic 90-degree rotations between both. Two other MAVs (1 and 2) are also present in the proximity, each MAV observing one of the buildings from a distance. For every image captured by these three MAVs, relative poses are computed between 1-0 and 2-0: and based on which estimate has a lower uncertainty, the final yaw angle of the rotating MAV is computed accordingly. In figure 5(left), we show a graph of the estimated yaw of MAV 0, where it can be seen that the estimated angle closely matches the ground truth. In figure 5(right), it can be noticed how MAV 1 or MAV 2 is chosen as the source of the relative yaw

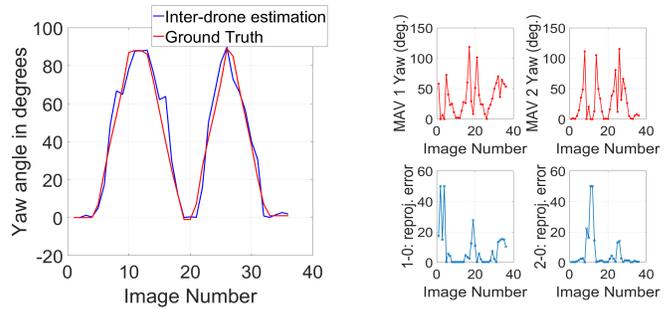

**Fig. 5:** Left: Yaw angle computed for a rotating MAV using information coming from two other MAVs. Right: Relative angles computed between 1-0 and 2-0, which are then used according to respective reprojection errors for the final estimate of the angle.

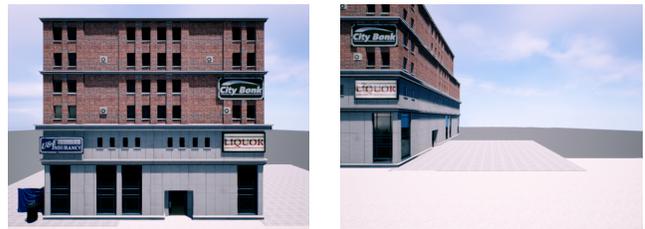

**Fig. 6:** Example images from the sparse environment used to test inter-MAV localization vs intra-MAV localization. One MAV observes a texture rich scene (left), whereas another can only observe a small part of it (right).

information based on the reprojection error (on which the uncertainty estimate depends).

We demonstrate the advantage of inter-MAV localization in another case involving a sparsely populated environment. Here, only one of the MAVs is able to observe a texture-rich part of the scene, and other MAVs can only be at a certain distance due to mission requirements, thus being unable to generate enough overlap between their observations and the global map. A scenario like this usually results in the PNP algorithm (intra-MAV localization) failing to generate an accurate pose, either due to a low number of inliers or a degenerate case because of bad locations of feature correspondences. In our simulation, one MAV observes a building directly, with two other MAVs located on either side with significantly low number of overlapping features, while navigating solely in the vertical direction (flying up and down). When the VCL algorithm is applied to this environment, we observe the higher robustness of the inter-MAV localization compared to the intra-MAV localization, as it is able to isolate new feature points that are visible between these views and use them to continue localizing.

Figure 6 shows sample images from the environment and the views of the 'host' MAV and a 'target' MAV, whose pose is to be estimated. In figure 7, it can be seen that the intra-MAV localization generates erratic poses for the target MAVs due to bad feature tracking, while the inter-MAV estimation performs better by isolating sufficient number of features that specifically belong to particular overlapping regions. Figure 7 shows the 3D plots of positions on the right, and a comparison of the profiles of the X positions of one of the target MAVs.

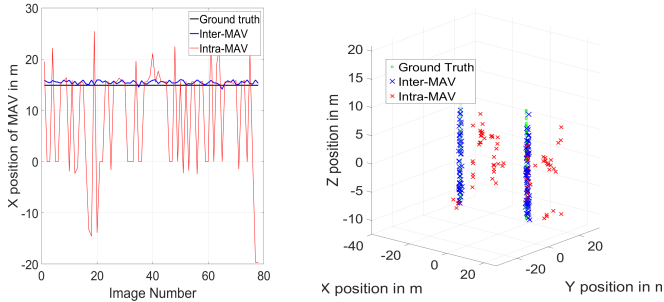
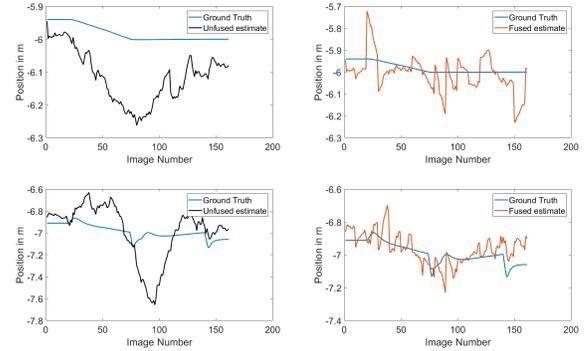

**Fig. 7:** Results from applying VCL to a sparsely populated environment and MAVs with low feature overlap. Left: Estimates of X position of an MAV: Inter-MAV estimation generates much more accurate poses than intra-MAV estimation. Right: 3D positions estimated using inter and intra-MAV localizations: intra-MAV localization displays a significantly higher error. Ground truth shown in green.

**Fig. 8:** Comparison of unfused estimates vs fused estimates and how they differ from the ground truth. Fusing relative with individual measurements (right) provides more corrections, periodically driving the estimation closer to the true value, thus resulting in a lower error overall.

### B. Fusion of inter and intra-MAV localization

In this section, we present results of fusion between intra-MAV and inter-MAV localization and attempt to demonstrate how occasional fusion with relative measurements helps increase the overall localization accuracy. For this test, we use a scenario with three MAVs flying a forward-backward trajectory while observing a common scene (a building). The MAVs are initially at locations close to this building, and collaborate at the first time step in order to generate a map, and as they are observing the same scene throughout, no map updates are triggered. But as the MAVs move backward in their trajectories and thereby away from the building, the accuracy of intra-MAV localization starts to suffer because of the increasing distance from the 3D points. To help with this, we trigger inter-MAV estimation between MAVs 0 and 1, and 0 and 2 every 10 images; and these relative estimates are fused with the individual estimates as described in section IV(E).

We observe a reduction in the RMS error compared with the ground truth when this fusion takes place; as opposed to when only intra-MAV localization takes place. Figure 8 shows a comparison of the profiles of a few unfused vs fused estimates of the trajectories, whereas figure 9 shows a 3D plot of the final estimates. As seen in figure 8, the relative measurements being more accurate, fusing these with the individual estimates forces the pose estimate periodically to be closer to the ground truth, whereas using only individual measurements causes it to drift when the MAVs are farther from the scene (middle part of the plot). In table 1, we show the RMS errors of the fused vs unfused estimates as compared to the ground truth, and it can be noted that fusion results in lower RMS errors and thus better localization. The total distance traveled by each MAVs was approximately 60m.

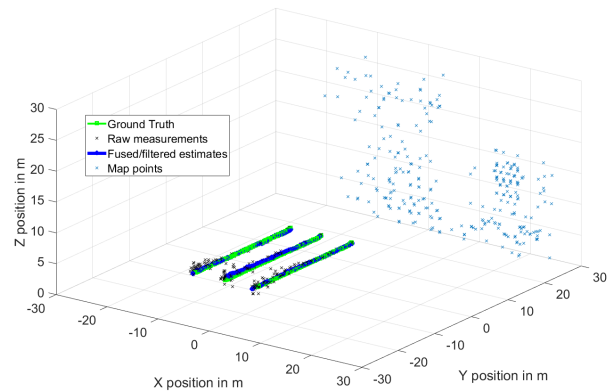

**Fig. 9:** 3D plots of ground truth, raw measurements and final fused estimates for a 3-MAV group executing a forward-backward trajectory. Map points are also shown in blue.

**TABLE I:** Comparison of localization accuracy between purely intra-MAV localization and inter-intra fused localization

| MAV ID | Axis | Fused | Unfused |
|---|---|---|---|
| 1 | X (cm) | 2.5 | 3.6 |
| | Y (cm) | 23.9 | 35.2 |
| | Z (cm) | 0.4 | 3.0 |
| 2 | X (cm) | 0.5 | 1.9 |
| | Y (cm) | 12.0 | 17.9 |
| | Z (cm) | 0.7 | 5.4 |

In figure 10, we display the 'covariances' obtained from the pose estimation parts of the algorithm. As these are a combination of solution uncertainty and reprojection errors, it can be seen that the intra-MAV measurements suffer in accuracy in the middle, corresponding to a large distance between the scene and the MAVs. In contrast, the inter-MAV measurements consistently maintain a comparatively better accuracy: which is reflected in the performance of the fusion

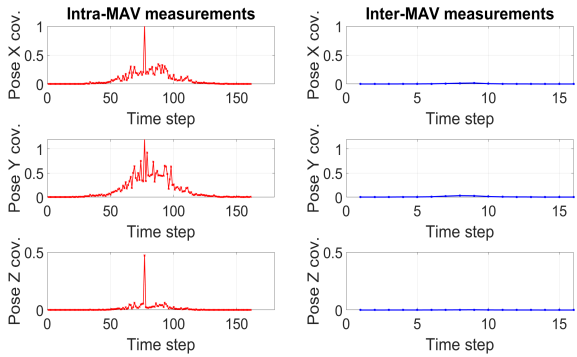

**Fig. 10:** Comparison of solution accuracy for inter-MAV localization and intra-MAV localization for the same trajectory displayed in figure 9. Inter-MAV localization exhibits higher confidence of solution throughout the trajectories.

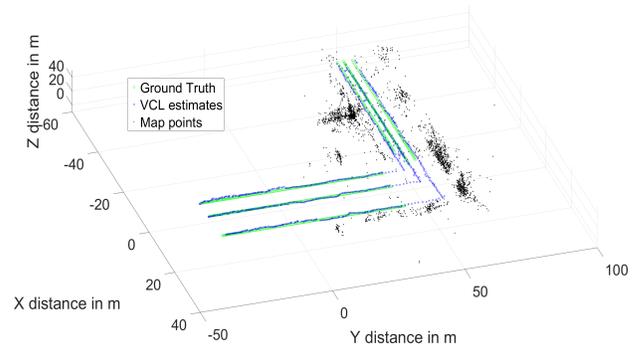

**Fig. 12:** Plots of ground truth (green) and VCL outputs (blue) for flights from the urban setting. Multiple reconstructions attempted during the flight resulted in map points shown here in black.

algorithm.

As a final test, we show localization for three MAVs flying through an urban environment, shown in figure 11. The MAVs start at the left part and travel to the top through a 90-degree turn. As this involves a considerable amount of distance, the localization algorithm performs intermediate map updates, as well as inter-MAV estimation during the turn as it involves pure rotation. Plots of ground truth and the estimates from the VCL algorithm are shown in figure 12.

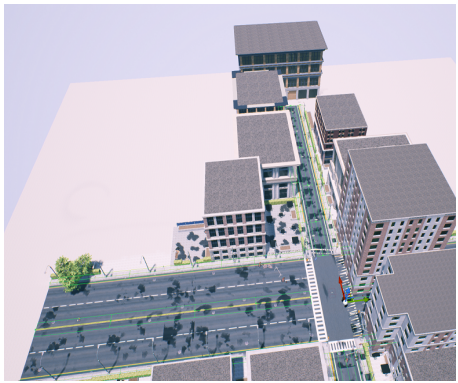

**Fig. 11:** Picture of an urban setting from Microsoft AirSim where three MAVs were flown.

## VI. CONCLUSION

In this paper, we present a vision based collaborative localization framework aimed at micro aerial vehicles equipped with forward facing monocular cameras. Feature detection and matching between the MAVs enables the creation of a 3D map that is then shared between them. Once a map is available, the MAVs are capable of alternating between intra-MAV localization: where each MAV attempts to track features from the map and estimate its own pose; and inter-MAV localization, where one MAV attempts to correct the pose of another MAV using a relative pose measurement. This algorithm was tested for image datasets coming from the MAV simulator Microsoft AirSim and preliminary results were obtained that show the advantages of collaboration during localization.

In terms of future work, we plan on testing it extensively on data from real flights. At the same time, this algorithm forms the base for our collaborative planning framework, where we attempt to perform uncertainty-aware path planning for swarms of micro aerial vehicles that are capable of localizing collaboratively. As far as further improvements go, the VCL algorithm currently works offline on pre-recorded data: a real time implementation could be possible while taking into account factors such as communication times, possibility of delay between measurements and so on. It is also possible to examine the possibility of MAVs splitting into multiple groups to visit different areas and coming together again: which would involve keeping a overlap graph in memory and ungrouping and regrouping the MAVs accordingly.